\documentclass[10pt,twocolumn,letterpaper]{article}

\usepackage[pagenumbers]{cvpr} % To force page numbers, e.g. for an arXiv version
\usepackage{graphicx}
\definecolor{cvprblue}{rgb}{0.21,0.49,0.74}
\usepackage[pagebackref,breaklinks,colorlinks,allcolors=cvprblue]{hyperref}
\def\paperID{*****} % *** Enter the Paper ID here
\def\confName{CVPR}
\def\confYear{2025}

\title{Exploring Temporal Dynamics in Event-based Eye Tracker}

\author{Hongwei Ren\thanks{Equal contribution, the order was determined by a random seed. \\ \dag corresponding author.} , Xiaopeng Lin\textsuperscript{*}, Hongxiang Huang\textsuperscript{*}, Yue Zhou, Bojun Cheng\dag\\
MICS Thrust\\
The Hong Kong University of Science and Technology (Guangzhou)\\
{\tt\small \{hren066, xlin746, hhuang516, yzhou883\}@connect.hkust-gz.edu.cn, bocheng@hkust-gz.edu.cn}
}

\begin{document}
\maketitle
\begin{abstract}
% The ABSTRACT is to be in fully justified italicized text, at the top of the left-hand column, below the author and affiliation information.
% Use the word ``Abstract'' as the title, in 12-point Times, boldface type, centered relative to the column, initially capitalized.
% The abstract is to be in 10-point, single-spaced type.
% Leave two blank lines after the Abstract, then begin the main text.
% Look at previous \confName abstracts to get a feel for style and length.
%第一句 eye tracking 背景。\textbf{Hongxiang}

%第二句 event camera。\textbf{Xiaopeng}

%第三句 temporal dynamic 重要性。\textbf{Hongwei}

Eye-tracking is a vital technology for human-computer interaction, especially in wearable devices such as AR, VR, and XR. The realization of high-speed and high-precision eye-tracking using frame-based image sensors is constrained by their limited temporal resolution, which impairs the accurate capture of rapid ocular dynamics, such as saccades and blinks.
Event cameras, inspired by biological vision systems, are capable of perceiving eye movements with extremely low power consumption and ultra-high temporal resolution.
This makes them a promising solution for achieving high-speed, high-precision tracking with rich temporal dynamics. In this paper, we propose TDTracker, an effective eye-tracking framework that captures rapid eye movements by thoroughly modeling temporal dynamics from both implicit and explicit perspectives. TDTracker utilizes 3D convolutional neural networks to capture implicit short-term temporal dynamics and employs a cascaded structure consisting of a Frequency-aware Module, GRU, and Mamba to extract explicit long-term temporal dynamics. Ultimately, a prediction heatmap is used for eye coordinate regression. Experimental results demonstrate that TDTracker achieves state-of-the-art (SOTA) performance on the synthetic SEET dataset and secured Third place in the CVPR event-based eye-tracking challenge 2025. Our code is available at \url{https://github.com/rhwxmx/TDTracker}.
\end{abstract}    
\section{Introduction}
\label{sec:intro}
%eye tacking 一段。\textbf{Xiaopeng}

Eye tracking technology is pivotal to the advancement of human-computer interaction, finding extensive application in augmented reality (AR), extended reality (XR), medical diagnostics, and psychological research \cite{jin2024eye}. It provides precise measurements of rapid ocular movements, enabling sophisticated, intuitive gaze-based interactions, notably in advanced wearable platforms\cite{ma2024generative}. However, the rapid velocities (exceeding 300°/s) \cite{konrad2020gaze} and substantial accelerations (up to 24,000°/s²) \cite{ramachandran2002encyclopedia} characteristic of human eye movements necessitate sampling rates at the kilo-hertz level. Realizing such high precision within wearable devices is technically challenging due to strict limitations regarding power consumption, latency, and data processing capacity\cite{yang2023three,cheng2024memristive}. Hence, the development of high-speed and high-precision eye-tracking systems represents an indispensable requirement for contemporary technological innovation.

Traditional frame-based eye-tracking systems exhibit tracking delays of 45 to 81 milliseconds \cite{stein2021comparison}, insufficient for capturing rapid eye movements accurately at kilo-hertz rates. Achieving these high sampling rates significantly increases power consumption, while the resulting large data volumes require substantial bandwidth and energy for real-time processing \cite{lin2024fapnet, chen20233et, ding2024facet}. Event cameras provide an effective solution by eliminating redundant information and focusing solely on the dynamic elements within the scene. 

Event cameras are a type of bio-inspired vision sensor that respond to local changes in illumination intensity exceeding a predefined threshold, offering several advantages over traditional frame-based cameras \cite{lichtsteiner2008128}. They feature low latency, high dynamic range, and asynchronous operation, making them highly suitable for applications involving rapid movements and varying lighting conditions \cite{rebecq2019high, ren2023spikepoint, ren2024rethinking}. Specifically, event cameras can capture brightness changes with microsecond precision and provide a high temporal resolution, which is crucial for eye tracking because even the fastest eye movements, such as saccades, can be accurately captured and tracked. These characteristics enable event cameras to deliver a sparse but informative data stream that is highly efficient for eye tracking. However, despite these advantages, current algorithms for eye tracking using event cameras still face several challenges. 
% One of the primary issues is the effective extraction and utilization of the multi-granularity temporal informatio inherent in event data.
One of the primary issues is the effective extraction and utilization of the temporal dynamics inherent in event data.

%temporal dynamic 一段。\textbf{Hongwei}
Temporal dynamics, the characteristic of how eye movements and states evolve over time, play a crucial role in the design of robust eye-tracking systems \cite{bagci2004eye}. The ability to accurately track eye coordinates depends not only on spatial information but also on the temporal changes that occur during eye movements. These dynamics, such as blinking, gaze shifts, reset, and saccade, introduce variability that can disrupt conventional tracking methods, making it essential for eye trackers to account for temporal factors to maintain continuous and accurate tracking \cite{wang2024event}. From another perspective, temporal dynamics enable the tracker to adapt to both short-term and long-term fluctuations in eye state. For instance, when an eye blinks or moves rapidly, the tracker must be able to predict the next gaze point without losing track of the eye. Failing to capture these temporal patterns can lead to tracking errors, poor user experience, and a drop in performance. Thus, effectively modeling temporal dynamics improves the robustness and reliability of eye trackers, allowing them to perform well under a wider range of conditions.

In this paper, we sufficiently explore temporal dynamics in an event-based eye tracker named TDTracker. TDTtrack is principally composed of two distinct components: implicit temporal dynamic (ITD) and explicit temporal dynamic (ETD). The ITD component implicitly extracts short-term temporal features by leveraging 3D convolution neural networks, effectively capturing the nuanced variations in temporal patterns over brief periods. On the other hand, the ETD component explicitly extracts long-term temporal features through the cascading structure of three advanced temporal models: Frequency-Aware Module, GRU, and Mamba. This cascading approach enables the model to capture more complex and sustained temporal dependencies, enhancing its overall performance in dynamic tracking tasks. Unlike most eye trackers that directly regress coordinates, TDTracer generates heatmaps and employs Kullback-Leibler (KL) divergence for training, enabling post-processing based on probability distributions. We conduct comprehensive validation on both synthetic and real datasets, TDTracker achieved state-of-the-art (SOTA) performance on the SEET dataset, with fewer floating point operations (FLOPs) than the previous SOTA, EventMamba. Additionally, TDTracker secured third place in the CVPR 2025 Event-Based Eye Tracking Challenge.

\section{Related work} 
\subsection{Eye Tracking} %xiaopeng

Traditional frame-based eye-tracking methods rely on frame-based cameras and are categorized into model-based and appearance-based approaches \cite{morimoto2005eye}. Model-based methods identify eye geometry and align key features with predefined models but often require manual calibration and struggle with variations in lighting and eye anatomy \cite{guestrin2006general, wang2017real, lai2014hybrid, mestre2018robust}. Appearance-based techniques utilize deep learning to analyze eye images, demanding significant computational resources and extensive training data \cite{lu2014adaptive, mazzeo2021deep, kim2019nvgaze}. Additionally, frame-based cameras generally operate at frequencies up to 200 Hz, pushing beyond this threshold significantly boosts power consumption, thus exceeding the energy limitations of mobile wearable systems \cite{ding2024facet}.

Event-based eye tracking methods utilize the intrinsic properties of event data to deliver high frame rates with minimal bandwidth, enhancing energy efficiency compared to conventional frame-based systems \cite{iddrisu2024event}. Recent advancements in event-based eye tracking demonstrate significant methodological evolution across multiple research teams. Li et al. \cite{li2023track} propose a three-channel temporal encoding framework coupled with a lightweight convolution neural network for low-latency pupil event prediction, establishing a new paradigm for real-time processing. Subsequently, Chen et al. \cite{chen20233et} implemented a temporal binning transformation of event data, developing a novel Cross-Bottleneck ConvLSTM (CB-ConvLSTM) architecture that outperforms conventional CNN models in spatiotemporal feature extraction. In parallel, Ryan et al. \cite{ryan2023real} innovatively integrated Gated Recurrent Units (GRUs) with an adapted YoloV3 framework, enabling robust eye tracking through voxel grid representations of asynchronous event streams. The 2024 AIS Challenge on Event-Based Eye Tracking \cite{wang2024event} catalyzed methodological innovation in this domain, with participants employing advanced architectures including ConvLSTMs and Mamba-based \cite{wang2024mambapupil} models for spatiotemporal processing of raw event data. Competitive solutions demonstrated sophisticated data conversion techniques ranging from dynamic binary encoding to neuromorphic point cloud representations \cite{lin2024fapnet}. Notably, recent publications \cite{bonazzi2024retina, jiang2024eye} demonstrate the efficacy of Spiking Neural Networks (SNNs) in extracting temporal features, thereby enhancing tracking precision while achieving lower computational overhead compared to conventional Artificial Neural Networks (ANNs).

\subsection{Time Series Model} %hongxiang
In event-based eye tracking, time series analysis is critical. Accurately extracting short-term and long-term, as well as local and global temporal information, contributes significantly to understanding users' fixation behaviors and visual attention patterns. Traditionally, recurrent neural networks (RNNs) \cite{elman1990finding} and their variants, such as Long Short-Term Memory (LSTM) \cite{hochreiter1997long} and Gated Recurrent Units (GRU) \cite{cho2014learning}, have been widely utilized in processing eye-tracking data. \cite{chen20233et} significantly increases the network sparsity and reduces the computational load while maintaining accuracy by introducing the change-based hidden state input. \cite{zemblys2019gazenet} employs GRU as the core component of its network architecture to model eye-movement event sequences and detect the onsets and offsets of eye-movement events, thereby classifying the eye-movement data.
However, these models may encounter problems like vanishing or exploding gradients when handling long sequences, negatively impacting model performance. 
\begin{figure*}[t]
\centerline{\includegraphics[width=0.85\textwidth]{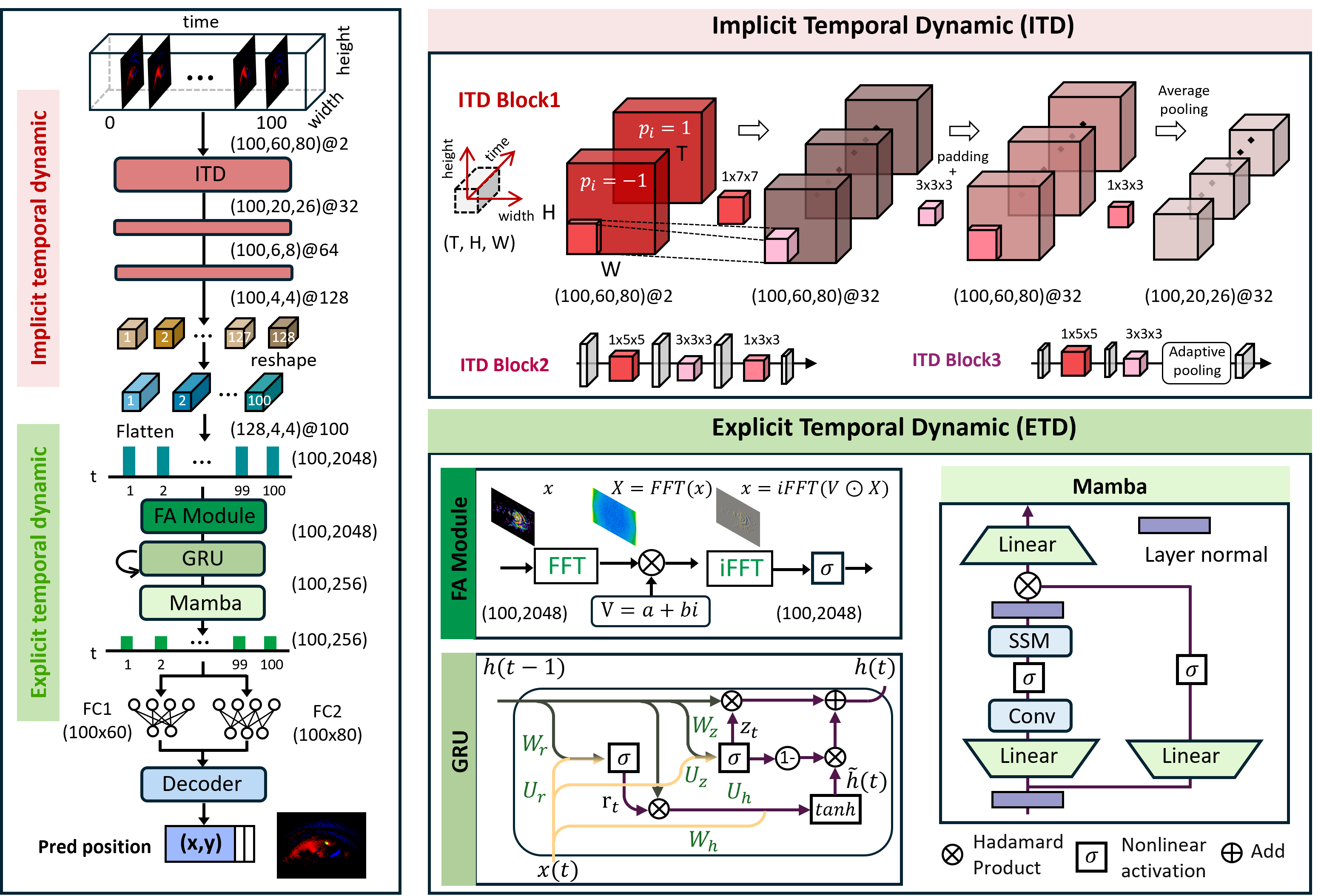}}
\caption{The architecture of TDTracker. TDTracker primarily comprises two components, Implicit Temporal Dynamic (ITD) and  Explicit Temporal Dynamic (ETD), with a structure featuring three ITD components to ensure effective feature abstraction. It employs a cascaded architecture of three distinct time series models to comprehensively capture temporal information.}
\label{fig:TDTracker}
\end{figure*}

Unlike RNNs, the Mamba model, as a selective state-space model, demonstrates unique advantages in time series modeling. By integrating the recursive characteristics of RNNs with the parallel computation abilities of CNNs, Mamba maintains linear complexity while enhancing information filtering capabilities, enabling efficient processing of long sequences and capturing global patterns. Currently, Mamba has been applied to various event-related tasks, \cite{ren2024rethinking} effectively enhances the extraction of explicit temporal features in event sequences by introducing the Mamba. \cite{wang2024mambapupil} models the hidden states of eye movement patterns in a dual recurrent module to selectively focus on valid eye motion phases, thereby improving the stability and accuracy of eye tracking.
However, the fixed hidden-state dimension in Mamba might be insufficient to represent extremely long sequences effectively. Regardless of whether using Mamba or RNNs, all these models may face the problem of information forgetting in long-term modeling, necessitating further improvements.

To better tackle the problem of long-term information forgetting, some studies have introduced the Fast Fourier Transform (FFT) \cite{cooley1965algorithm}. Based on the principle that multiplication in the frequency domain is equivalent to convolution in the time domain, FFT enables models to learn holistic information across the entire time domain and helps mitigate information forgetting in subsequent temporal modeling tasks. Currently, FFT has been employed in some tasks, such as super-resolution \cite{chen2024large, li2018frequency}, low-level tasks \cite{kim2024frequency, mao2023intriguing}, and human pose estimation \cite{zhao2023poseformerv2}.
However, no research has yet applied FFT methods to temporal information extraction in eye-tracking tasks, presenting a novel direction for future research.

\section{Method}
In this section, we provide a detailed explanation of TDTracker, covering the representations of raw events, the two key modules for implicit temporal dynamic and explicit temporal dynamic capturing within the network architecture, as well as the loss function specifically designed for heatmap prediction.
\subsection{Event representations}
Event cameras produce raw events that capture changes in the environment's illumination, encoding information in terms of 2 spatial dimensions, 1 temporal dimension, and 1 polarity (2S-1T-1P). These raw events, denoted as $\mathcal{E}$, can be formally expressed as:
\begin{equation}
        \mathcal{E} = \left\{e_i=(x_i,y_i,t_i,p_i)\right\}, i \in [1,2 \ldots, n],
\end{equation}
In this formulation, $(x, y)$ specifies the spatial location of the event emission, $t$ refers to the timestamp, $p$ denotes the polarity, and $i$ indicates the index of the $i_{\text{th}}$ element within the event stream.

Numerous approaches involve converting raw events into an event frame, which serves as a static representation of the dynamic event stream\cite{shariff2024event}. This frame is created by aggregating the number of events or their polarity at each pixel location as the following equation.
\begin{align}
\mathcal{F}(x, y) = \sum_{t_i \in (t_0, t_n)} p_i\text{ or } |p_i|\mid (t_i,p_i) \in e_i ,
\end{align}
Here, \( x \) and \( y \) denote the spatial coordinates within the event frame, where \( x \) ranges from 1 to \( w \) and \( y \) ranges from 1 to \( h \), with \( w \) and \( h \) signifying the width and height of the frame resolution, respectively. The event frame encapsulates the collected event data over discrete time intervals, effectively summarizing the temporal evolution of the events. This aggregated information is crucial for subsequent processing steps or analytical procedures, providing a comprehensive representation of the event dynamics across the defined spatial grid.

Another widely used representation is the Voxel, denoted as \( \mathcal{V} \), which is defined by the following set of equations\cite{deng2022voxel}. The time axis is divided into discrete intervals, referred to as time bins, with the boundaries of these bins specified by \( l_k \), where \( k \in [0, b] \) corresponds to the index of the \( k_{\text{th}} \) time bin. In particular, the boundary \( l_k \) is computed using the following formula:
\begin{align}
    l_k = t_0 + k\cdot \frac{t_n - t_0}{b},\quad k \in [0,b],
\end{align}
where \( t_0 \) and \( t_n \) represent the start and end times of the event sequence, respectively, and \( b \) denotes the total number of bins. The voxel representation \( \mathcal{V} \) is constructed by aggregating the event data \( e_i = \{(t_i, p_i)\} \), where each event \( e_i \) consists of a timestamp \( t_i \) and a polarity \( p_i \). These events are grouped into respective time bins \( (l_k, l_{k+1}) \), with \( l_k \) and \( l_{k+1} \) being the boundaries of the \( k_{\text{th}} \) time bin, as follows:
\begin{align}
    \mathcal{V}(x, y, k) =  \sum_{t_i \in (l_k, l_{k+1})} p_i \text{ or } |p_i|\mid (t_i,p_i) \in e_i,
\end{align}

Furthermore, the Binary Map representation \( \mathcal{B} \) combines \( b \) binary frames to generate a sequence of \( b \) bits for each pixel \cite{barchid2022bina}. For each pixel, the Binary Map considers this sequence of \( b \) bits as a representation in a different number system. For instance, a sequence of eight binary frames can be encoded as a single 8-bit unsigned integer frame. Additionally, the Binary Map approach can be extended in a manner similar to binary event frames, producing a sequence of \( T \) frames, each represented as \( b \)-bit numbers. This approach indirectly yields a sparse representation, relying on \( T \times b \) binary event images.
\subsection{Implicit Temporal Dynamic (ITD)}
TDTracker leverages 3D convolution applied to the frame-based representation over the dimensions \( (T, H, W) \), allowing the model to implicitly abstract the temporal dimension \( T \) into spatio-temporal features that effectively capture both temporal and spatial patterns within the data. The feature extraction process is structured into three sequential stages, each designed to progressively increase the receptive field as shown in \cref{fig:TDTracker}. Our method chooses the Binary Map representation as the preferred form of representation.
% This increase enables the model to capture progressively larger contexts and more intricate dependencies across both spatial and temporal dimensions. 
% To address challenges such as overfitting and the rapid explosion of model parameters, TDTracker incorporates a downsampling operation that reduces the spatial dimensions \( H \) and \( W \) in each stage. This operation serves multiple purposes: it alleviates the computational burden, reduces memory usage, and ensures the model generalizes effectively by preventing overfitting. As a result, TDTracker maintains a balance between model learning capacity and computational efficiency.

The input to the TDTracker is represented as the frame-based tensor \( \mathbf{I} \in \mathbb{R}^{C \times T \times H \times W} \), where \( T \) denotes the temporal dimension, and \( H \) and \( W \) represent the spatial dimensions. To begin the feature extraction process, TDTracker applies a 3D convolution operation on the input to capture spatial features using a kernel of size \( (1, K_s, K_s) \), where \( K_s \) represents the spatial kernel size. This convolution operation can be expressed mathematically as:
\begin{align}
\mathbf{F}_S = \text{Conv3D}(\mathbf{I}; \mathbf{W}_S) + \mathbf{B}_S
\quad \mathbf{W}_S \in \mathbb{R}^{ 1 \times K_s \times K_s}
\end{align}
where \( \mathbf{W}_S \) is the convolution kernel, \( \mathbf{B}_S \) is the corresponding bias term, and \( \text{Conv3D} \) denotes the 3D convolution operation performed over the spatial dimensions \( H \) and \( W \). 

Following this, TDTracker applies another 3D convolution operation on the intermediate feature map \( \mathbf{F}_S \) to extract temporal features using a kernel of size \( (K_t, K_t, K_t) \), where \( K_t \) is the temporal kernel size. We call this convolution named \textbf{Implicit-conv}. This operation abstracts the temporal dimension \( T \) into spatio-temporal features, capturing dynamic changes in the temporal domain as well as spatial patterns within the events:
\begin{align}
\mathbf{F}_T = \text{Conv3D}(\mathbf{F}_S; \mathbf{W}_T) \quad
\mathbf{W}_T \in \mathbb{R}^{ K_t \times K_t \times K_t}
\end{align}
This step allows the model to encode the implicit temporal dynamics alongside the spatial characteristics of the data.

As the network deepens, TDTracker progressively increases the receptive field at each stage. This increase enables the model to better capture larger contexts and more complex dependencies within the data. In order to mitigate the risk of overfitting and control the rapid growth of parameters, TDTracker employs average downsampling on the feature maps in the spatial dimensions \( H \) and \( W \) at each stage. This downsampling operation, typically implemented as pooling, reduces the spatial dimensions while preserving essential features. The downsampling strategy not only minimizes the computational cost but also aids in the preservation of critical spatial information, thus improving the model's generalization capability.

After passing through multiple stages of convolution and downsampling, the final feature map produced by ITD effectively captures the relevant implicit temporal and spatial patterns within the data, providing the model with the necessary features to make accurate predictions.

\subsection{Explicit Temporal Dynamic (ETD)}
TDTracker utilizes a combination of three distinct types of time-series models stacked together to effectively extract explicit temporal features from the sequences. They are Frequency-aware Module, GRU, and Mamba. The following section describes them using mathematical methods.
\subsubsection{Frequency-aware Module}
One-dimensional discrete Fourier Transform (DFT) is employed to convert the features to the frequency domain by the following formulation:
\begin{equation}
X[k]=\sum_{n=0}^{T-1} x[n] e^{-j \frac{2 \pi}{T} k n}, \quad k=0,1, \ldots, T-1
\end{equation}
where $j$ is the imaginary unit, $x$ represents the signals in the temporal domain, $X$ denotes the spectrum at different frequencies, and $T$ is the length of temporal signals $x$. Additionally, the inverse DFT can recover spectrum to temporal signals by the following formulation:
\begin{equation}
x[n]=\frac{1}{T}\sum_{k=0}^{T-1} X[k] e^{j \frac{2 \pi}{T} k n}, \quad n=0,1, \ldots, T-1
\end{equation}
Mathematically, the $X[T-k] = X^{*}[k]$, where $k\in[0,\frac{T}{2}]$, it means the specturm $X$ is conjugate symmetric. Therefore, the transformed frequency domain spectrum only needs $\frac{T}{2}+1$ long enough to be recovered to the original signal.

Once the spectrum is obtained through DFT, we can initialize a learnable filter $V$ with dimensionality matching that of the spectrum and perform the Hadamard product between the spectrum and the filter. 
% \begin{equation}
%     \hat{X} = V\odot X
% \end{equation}
In summary, the specific process of the frequency-aware module is as follows:
\begin{equation}
\begin{aligned}
    X &=  \text{FFT}(x),\\
    \hat{X} &=   V\odot X,\\
    x = \text{iFFT}&(\hat{X}),x = \sigma(x),
    \label{eq: fft_process}
\end{aligned}
\end{equation}
where $\sigma$ is a nonlinear activation function and $\odot$ means Hadamard product.
\subsubsection{Gated Recurrent Unit}
GRU (Gated Recurrent Unit) block is designed to effectively capture sequential dependencies by incorporating gating mechanisms that control the flow of information. This model is particularly adept at learning long-term temporal patterns, making it suitable for time-series data with intricate relationships over time. The GRU processes inputs and updates the hidden states as follows:
\begin{align}
r_t &= \sigma(W_r x_t + U_r h_{t-1} + b_r) \\
z_t &= \sigma(W_z x_t + U_z h_{t-1} + b_z) \\
\tilde{h}_t &= \tanh(W_h x_t + U_h (r_t \odot h_{t-1}) + b_h) \\
h_t &= (1 - z_t) \odot \tilde{h}_t + z_t \odot h_{t-1}
\end{align}
\( r_t \) is the reset gate, \( z_t \) is the update gate, \( \tilde{h}_t \) is the candidate hidden state, \( h_t \) is the current hidden state, \( x_t \) is the input at time step \( t \), and \( h_{t-1} \) is the hidden state from the previous time step. \( \sigma \) denotes the nonlinear activation function, \( \tanh \) is the hyperbolic tangent activation function, and \( \odot \) represents element-wise multiplication.
\begin{table*}[t]
\centering
\renewcommand\arraystretch{1.2}
\scalebox{0.8}{
\begin{tabular}{ccccccccc}
\hline
Method          & Resolution & Representation & Param (M)         & FLOPs (M)      & $p_{3}$             & $p_{5}$             & $p_{10}$            & mse(px)       \\ \hline

\textbf{TDTracker} &60 x 80 &F &3.248 &318 &0.953 & 0.996& 1.0& 1.30\\ \hline
MambaPupil \cite{wang2024mambapupil} & 60 x 80  & F     & 8.608 & 149 & 0.905 & 0.995 & 1.0 & 1.64 \\

EventMamba\cite{ren2024rethinking} & 180 x 240  & P     & 0.903 & 476 & 0.944 & 0.992 & 0.996 & 1.48 \\
FAPNet \cite{lin2024fapnet}& 180 x 240  & P     & 0.29 & 58.7 & 0.920 &0.991 & 0.996 & 1.56 \\
PEPNet\cite{ren2024simple}          & 180 x 240  & P              & 0.64          & 443           & 0.918          & 0.987          & 0.998          & 1.57         \\
$\text{PEPNet}_{\text{tiny}}$\cite{ren2024simple}    & 180 x 240  & P              & 0.054        & 16.25       & 0.786          & 0.945          & 0.995          & 2.2           \\
$\text{PointMLP}_{\text{elite}}$\cite{ma2022rethinking} & 180 x 240  & P              & 0.68          & 924           & 0.840          & 0.977          & 0.997         & 1.96          \\
PointNet++\cite{qi2017pointnet++}      & 180 x 240  & P              & 1.46          & 1099          & 0.607          & 0.866          & 0.988          & 3.02          \\
PointNet\cite{qi2017pointnet}        & 180 x 240  & P              & 3.46          & 450           & 0.322          & 0.596          & 0.896          & 5.18          \\
CNN\cite{chen20233et}              & 60 x 80    & F              & 0.40          & 18.4         & 0.578          & 0.774          & 0.914          & -             \\
TENNs \cite{pei2024lightweight}& 60 x 80  & F     & 0.649 & 26 & 0.765 & 0.937 & 0.992 & 2.37 \\
CB-ConvLSTM\cite{chen20233et}    & 60 x 80    & F              & 0.42          & 18.68         & 0.889          & 0.971          & 0.995          & -             \\
ConvLSTM\cite{chen20233et}         & 60 x 80    & F              & 0.42          & 42.61         & 0.887          & 0.971          & 0.994          & -             \\ \hline
\end{tabular}
}
\caption{The results of TDTracker on synthetic SEET dataset.}
\label{table: SEET}
\end{table*}

\subsubsection{Mamba}
This block mainly integrates the sixth version of the State Space Model (SSM), which is well able to parallel and focus on long time series of information, such as temporal correlations between 100 sequences. The SSM extracts explicit temporal features that can be expressed as follows:
\begin{align}
    h_t = & \bar{\mathbf{A}}h_{t-1} + \bar{\mathbf{B}}x_t,  \quad t \in [1,T') \quad h\in D'\\
    y_t = & \mathbf{C}h_t, \quad y\in D'\\
    \bar{\mathbf{A}} = & exp(\Delta \mathbf{A}), \quad \bar{\mathbf{A}},\bar{\mathbf{B}}:(T',D',D')\\
    \bar{\mathbf{B}}=&(\Delta \mathbf{A})^{-1}(\exp (\Delta \mathbf{A})-I) \cdot \Delta \mathbf{B},  
\end{align}
Where $x_t,h_t,$ and $y_t$ are the SSM's discrete inputs, states, and outputs. $\mathbf{A}$, $\mathbf{B}$ and $\mathbf{C}$ are the continuous system parameters, while $\bar{\mathbf{A}}$, $\bar{\mathbf{B}}$ and $\Delta$ are the parameters in the discrete system by the zero-order hold rule. $T'$ and $D'$ are the number and dimension of events for the current stage, respectively.
The whole $GlobalFE$ Extractor can be represented by this formula:
\begin{align}
    \mathcal{ST} = & \quad \text{Mamba}(\mathcal{SA}) \\
    \mathcal{ST} = & \quad \text{ResB}(\mathcal{ST})
\end{align}
Where ResB is a residual block, and Mamba extracts explicit temporal features in the $T'$ dimension, while ResB further abstracts the spatial and temporal. 
\subsection{Loss Function}
\begin{figure}[t]
\centerline{\includegraphics[width= 8 cm]{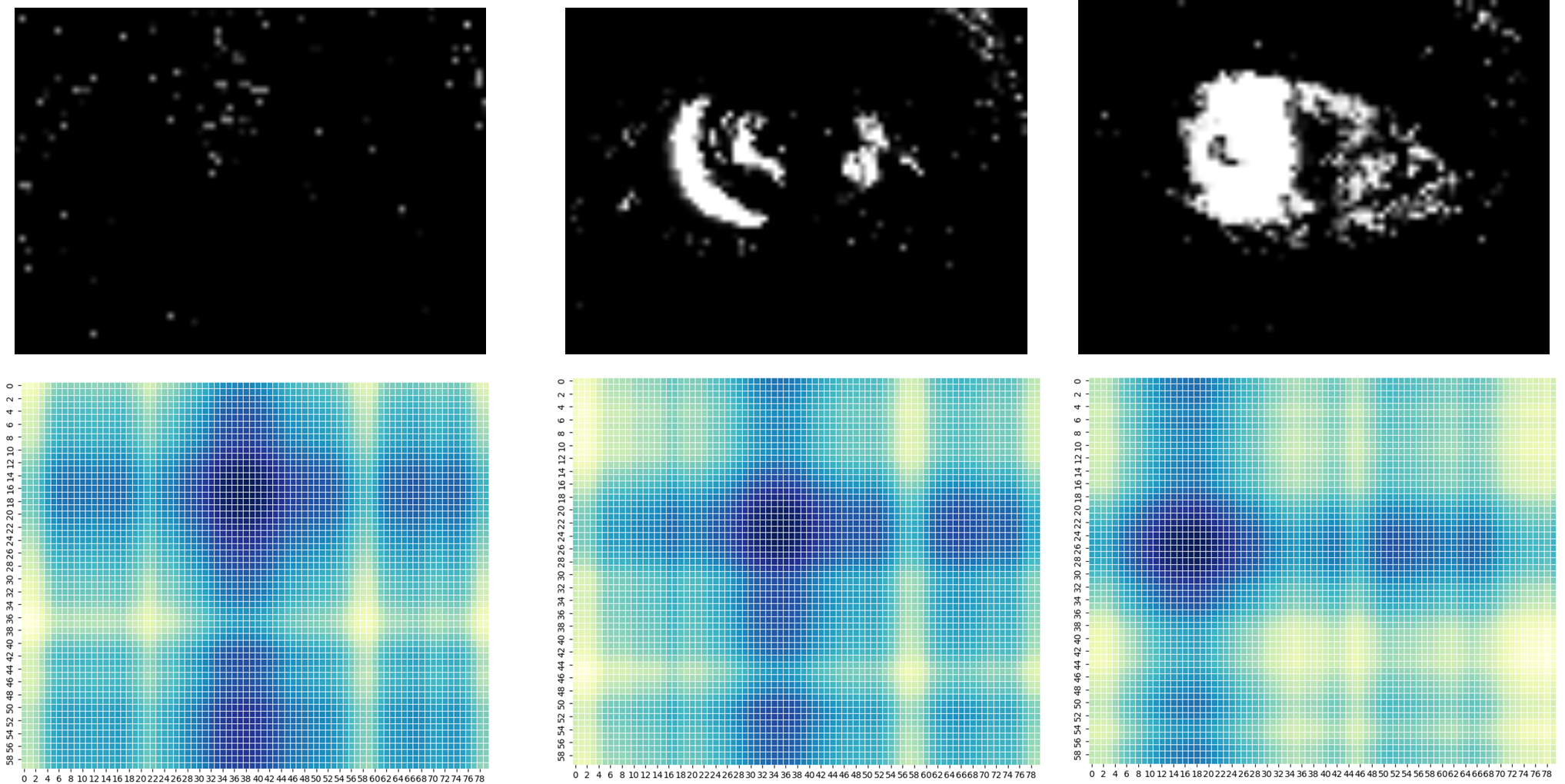}}
\caption{The visualization heatmap generated by the TDTracker.}
\label{fig: heat}
\end{figure}
In several works that utilized RGB images with a CNN model \cite{newell2016stacked,wei2016convolutional}, it was shown that transforming the labels into a 2D heatmap and predicting the heatmap provides a more effective approach in regression tasks. We transform the 2D labels $(x, y)$ into two 1D heat vectors $(\text{X}, \text{Y})$, corresponding to the evaluation metric resolution size.
\begin{align}
    \text{X} &= [v_0,v_1,v_2,\cdots v_W], v_x = 1\\
    \text{Y} &= [v_0,v_1,v_2.\cdots v_H], v_y = 1 
\end{align}
These vectors are one-hot encoded and then blurred using a Gaussian kernel, with shapes of (H, 1) and (W, 1), respectively.
\begin{align}
    v_{i}=\frac{1}{\sqrt{2 \pi} \sigma} \exp \left(-\frac{\left(i-\hat{v} \right)^{2}}{2 \sigma^{2}}\right) \quad \hat{v} = x\text{ or }y,
\end{align}
We utilize the KL divergence loss to measure the difference between two probability distributions, P and Q. It is commonly used as a loss function in various machine learning models to quantify how one probability distribution diverges from a second reference distribution.
\begin{align}
    L_{\text{KL}}(P \parallel Q) = \sum_{i} P(i) \log \left( \frac{P(i)}{Q(i)} \right)
\end{align}
Where $P(i)$ represents the probability of index i under the distribution $P$, while $Q(i)$ represents the probability of index i under the distribution $Q$. The summation runs over all possible indices. So, the final loss is obtained from the predicted $\hat{\text{X}}$ and $\hat{\text{Y}}$ and the KL scatter of the true labels X and $Y$.
\begin{align}
    L_{total} = L_{\text{KL}}(\text{X} \parallel \hat{\text{X}}) + L_{\text{KL}}(\text{Y} \parallel \hat{\text{Y}})
\end{align}
The final output is the coordinates with the highest probability on $\hat{\text{X}}$ and $\hat{\text{Y}}$, and we visualized the heatmap as shown in the \cref{fig: heat}.

\section{Experiment}
\begin{figure*}[ht]
\centerline{\includegraphics[width=0.75\textwidth]{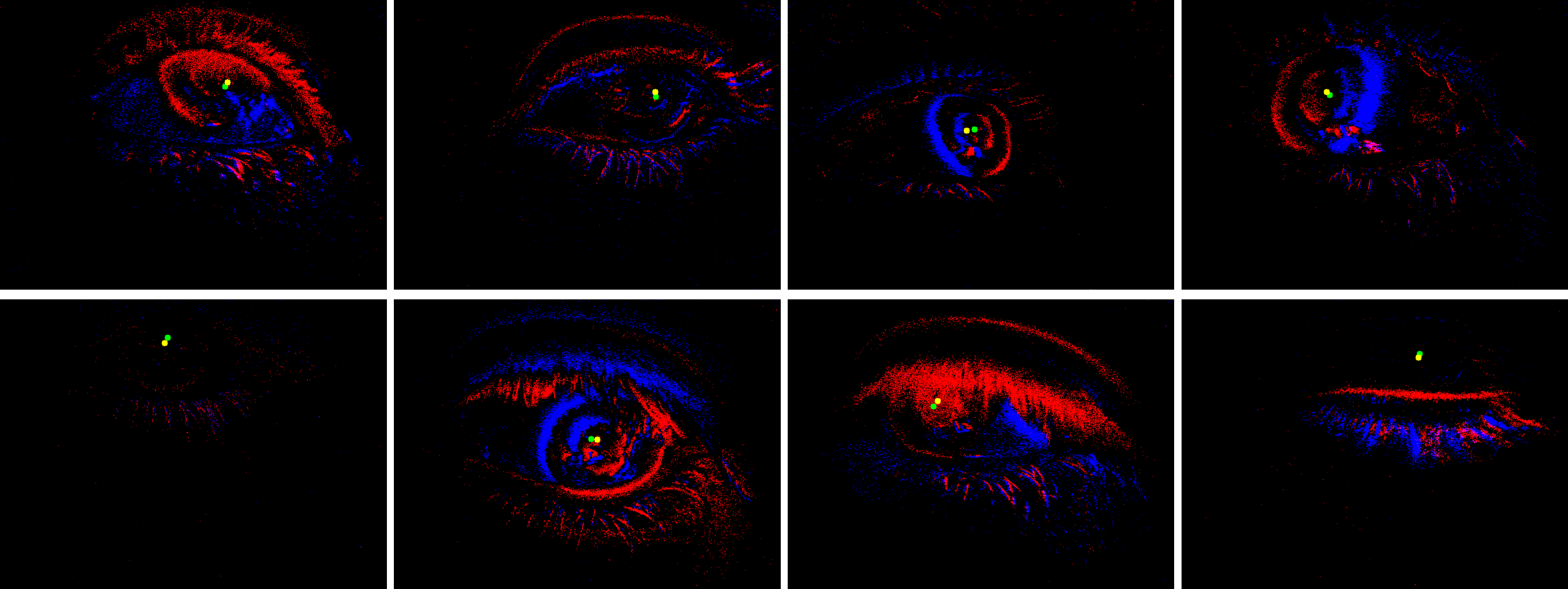}}
\caption{The visualization results of the TDTracker. The green dot in the figure stands for the Ground Truth label and the yellow dot is the prediction results genetated by TDTracker.}
\label{fig:result}
\end{figure*}

\begin{figure*}[ht]
\centerline{\includegraphics[width= 14 cm]{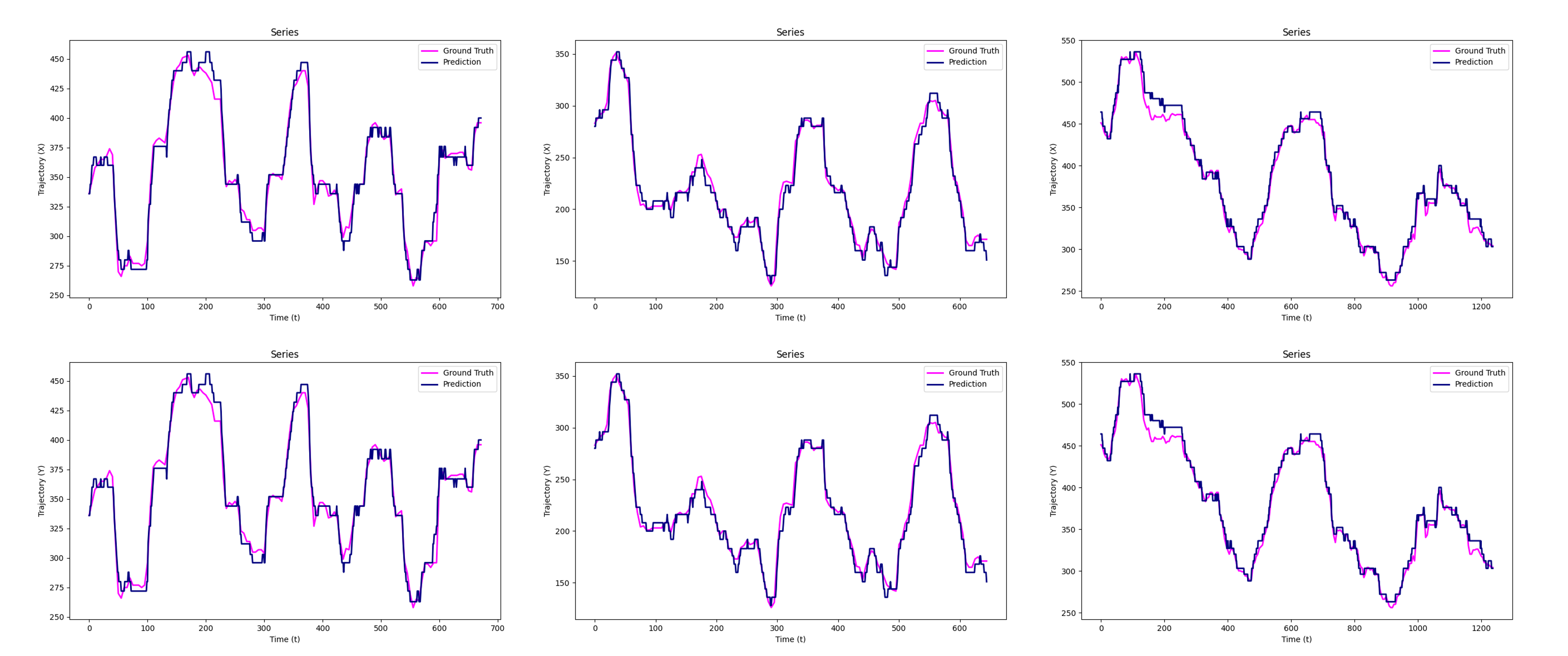}}
\caption{The visualization trajectory comparison between the Ground Truth label and the prediction results gererated by the TDTracker.}
\label{fig:trajectory}
\end{figure*}

In the experiments, the proposed method is evaluated using two datasets: a real-world event dataset 3ET+ 2025 provided by the challenge and a synthetic event dataset SEET. The effectiveness of the eye-tracking method is assessed by measuring the accuracy of pupil location prediction. The Euclidean distance between the predicted pupil location and the ground truth label serves as the evaluation metric. Tracking success is defined as an instance where the distance error falls within p pixels, which is subsequently used to calculate the tracking rate.

\subsection{Dataset}
The SEET dataset is a synthetic eye-tracking dataset created from an RGB dataset called Labeled Pupils in the Wild \cite{chen20233et}. Event streams are generated using the v2e simulator \cite{hu2021v2e}at a resolution of 240 × 180, with a time step $\Delta T$ of 4.4 ms, ensuring synchronization with the frame rate of the original RGB dataset.

The 3ET+ dataset \cite{wang2024event} is an event-based eye-tracking dataset that features real-world events captured using the DVXplorer Mini camera. It includes data from 13 participants, each involved in 2 to 6 recording sessions. During these sessions, participants engage in five types of activities: random movements, saccades, reading text, smooth pursuit, and blinking. The total data size is 9.2 GB. Ground truth labels are provided at 100Hz.
\subsection{Implement Details}
Our server leverages the PyTorch deep learning framework and selects the AdamW optimizer with an initial learning rate set to $2\cdot e^{-3}$, which employs a Cosine decay strategy, accompanied by a weight decay parameter of $1\cdot e^{-4}$. This configuration is meticulously chosen to enhance the model's convergence and performance through adaptive learning rate adjustments. Training is conducted on an NVIDIA GeForce RTX 4090 GPU with 24GB of memory, enabling a batch size of 16.
\begin{table*}[tb]
\centering
\renewcommand\arraystretch{1.2}
\scalebox{0.8}{
\begin{tabular}{ccccccccc}
\hline
Method          & Resolution & Representation & Param (M)         & FLOPs (M)      & $p_{3}$             & $p_{5}$             & $p_{10}$            & mse(px)       \\ \hline

\textbf{TDTracker} &60 x 80 &F &3.248 &318 &0.912 & 0.972& 0.992& 1.62\\
Mambapupil &60 x 80 &F &8.608 &149 &0.884 &0.972 &0.993 &1.77 \\
TENNs &60 x 80 &F &0.649 &26 &0.809 & 0.889& 0.995& 2.56 \\ \hline

\end{tabular}
}
\caption{The results of TDTracker on 3ET+ 2025 dataset. The ground truth is obtained from the interpolation of 3ET+ 2024. The result of the comparison model is an open-source code reproduction version.}
\label{table: 3ET+}
\end{table*}
\subsection{Results on SEET Dataset}
We present the results of TDTracker on the synthetic SEET dataset, as shown in \cref{table: SEET}. TDTracker is designed with a resolution of $60 \times 80$, utilizing a frame-based representation. The model achieves a parameter of 3.248 million and requires 318 million FLOPs. In terms of tracking accuracy, TDTracker demonstrates superior performance with a high $p_{3}$ score of 0.953, $p_{5}$ of 0.996, and $p_{10}$ of 1.0, which indicates its robust capability in maintaining precise tracking over time. Furthermore, the model achieves an MSE of 1.30 pixels, highlighting its accuracy in localizing the eye within the event stream. In comparison with other state-of-the-art methods, TDTracker performs favorably, especially when considering the trade-off between accuracy and computational efficiency. For instance, MambaPupil \cite{wang2024mambapupil} not only yields a lower $p_{3}$ of 0.905 but also has a substantially larger number of parameters (8.608 million), resulting in a higher MSE of 1.64 pixels. Additionally, methods such as EventMamba \cite{ren2024rethinking} report better accuracy but with significantly more computational complexity (476 million FLOPs). The results demonstrate that TDTracker achieves a strong balance between model efficiency and accuracy, making it a promising candidate for real-time eye-tracking tasks in resource-constrained environments.
\begin{table*}[tb]
\centering
\renewcommand\arraystretch{1.2}
\scalebox{0.8}{
\begin{tabular}{cccccccc}
\hline
Method           & Param (M) & FLOPs         & $p_{1}$      & $p_{3}$             & $p_{5}$             & $p_{10}$            & mse(px)       \\ \hline

w all  &3.248 &318 &0.391 &0.912 & 0.972& 0.992& 1.62\\
w/o FFT  &3.04 &265 &0.383 &0.921 & 0.973& 0.992& 1.63\\
w/o Mamba   &2.81 &274 &0.371 &0.903 & 0.967& 0.985& 1.70 \\ 
w/o Implict-Conv  &2.458 &53 &0.178 &0.783 &0.944 & 0.989& 2.32 \\ 
\hline

\end{tabular}
}
\caption{Abalation study of TDTracker on 3ET+ 2025 dataset. The ground truth is obtained from the interpolation of 3ET+ 2024.}
\label{table: ablation key}
\end{table*}
\begin{table}[tb]
\centering
\renewcommand\arraystretch{1.2}
\scalebox{0.8}{
\begin{tabular}{cccccc}
\hline
Representation         & $p_{1}$   & $p_{3}$             & $p_{5}$             & $p_{10}$            & mse(px)       \\ \hline

Event Frame &0.237 &0.849 &0.953  &0.992 &2.06 \\
Voxel &0.230 &0.819 &0.952  &0.993 &2.10 \\
Binary Map &0.265 &0.862 &0.962  &0.993 &1.89 \\
 \hline

\end{tabular}
}
\caption{The ablation of different representation on 3ET+ 2025 dataset. No augment method is employed.}
\label{table: ablation representation}
\vspace{-0.5cm}
\end{table}
\subsection{Results on 3ET+ 2025 Dataset}
The performance of TDTracker on the 3ET+ 2025 dataset is presented in \cref{table: 3ET+}. TDTracker achieves strong results with $p_{3}$ score of 0.912, $p_{5}$ of 0.972, and $p_{10}$ of 0.992, and a MSE of 1.62 pixels. Compared to MambaPupil, which has a higher parameter count (8.608M), TDTracker performs more efficiently with lower computational cost and better tracking accuracy (MSE of 1.77 pixels). While TENNs is more computationally efficient, its tracking accuracy is lower, with a $p_{3}$ of 0.809 and an MSE of 2.56 pixels. Overall, TDTracker strikes a good balance between efficiency and accuracy. What's more, TDTracker’s inference time is 1.7923 ms on RTX 4090, and we draw serval the predict coordinates and trajectory of TDTracker in \cref{fig:result,fig:trajectory}.

\subsection{Ablation Study}
\subsubsection{Key Module Ablation}
The ablation study of the key module for TDTracker on the 3ET+ 2025 dataset, shown in \cref{table: ablation key}, demonstrates the impact of different modules on performance. The full model denoted as "w all," achieves the best performance of 1.62 pixels. Removing the FFT module ("w/o FFT") results in a slight drop in performance, particularly at $p_{1}$, while still maintaining strong precision scores ($p_{3} = 0.921$, $p_{5} = 0.973$). Excluding the Mamba ("w/o Mamba") further reduces accuracy, especially at $p_{1}$, and leads to an MSE of 1.70 pixels. The most significant performance degradation occurs when the Implicit-conv module is removed ("w/o Implicit-Conv"), where the precision scores drop drastically, and the MSE increases to 2.32 pixels. These results highlight the critical role of each component in enhancing the tracking accuracy and efficiency of TDTracker.
\subsubsection{Representation Ablation}
The ablation study on different representations for TDTracker shows that the choice of representation significantly affects the model's performance in \cref{table: ablation representation}. Among the tested options, the Binary Map representation consistently outperforms the others, offering the best balance of accuracy and precision. While the Event Frame and Voxel representations also yield strong results, they tend to have slightly lower performance, particularly in terms of precision and error. The Binary Map representation demonstrates the most robust tracking performance, suggesting its superior ability to capture relevant features for TDTracker. Overall, these findings highlight the importance of selecting an appropriate representation to optimize tracking accuracy.

\subsection{Challenge Post-process}
In the competition, we found that using 100 sequence training, 200 sequence testing had the highest accuracy (MSE: 1.62 to 1.55). However, since the parameters of the Frequency-aware module are tied to the sequence length, we canceled this module in the competition. In addition, since our model does not consider open and closed eye cases, we simply use the ratio of the number of up and down events as the basis for judgement (set to 0.09), and when the current ratio is smaller than this value, the inference eye coordinate of the changed sample is overwritten by the inference value of the closest to this sample. What's more, we differ from directly regressing coordinate information by using a predicted probability density map, which provides an additional probability of the model predicting this image as shown in \cref{fig: heat}. If the probability is less than 0.5, we do not believe the predicted result. After post-processing, the MSE is optimized to 1.4936.

\section{Conclusion}
In this paper, we present TDTracker, an event-based eye tracker that effectively captures both implicit and explicit temporal dynamics. By leveraging 3D CNNs and a cascaded architecture of Frequency-Aware Module, GRU, and Mamba models, TDTracker improves tracking performance and enhances robustness. Experiments demonstrate SOTA's performance on the SEET dataset and third place in the CVPR 2025 Event-Based Eye-tracking Challenge.

{
    \small
    \bibliographystyle{ieeenat_fullname}
    \bibliography{main}

\begin{thebibliography}{51}
\providecommand{\natexlab}[1]{#1}
\providecommand{\url}[1]{\texttt{#1}}
\expandafter\ifx\csname urlstyle\endcsname\relax
  \providecommand{\doi}[1]{doi: #1}\else
  \providecommand{\doi}{doi: \begingroup \urlstyle{rm}\Url}\fi

\bibitem[Bagci et~al.(2004)Bagci, Ansari, Khokhar, and Cetin]{bagci2004eye}
A~Murat Bagci, Rashid Ansari, A Khokhar, and E Cetin.
\newblock Eye tracking using markov models.
\newblock In \emph{Proceedings of the 17th International Conference on Pattern Recognition, 2004. ICPR 2004.}, pages 818--821. IEEE, 2004.

\bibitem[Barchid et~al.(2022)Barchid, Mennesson, and Dj{\'e}raba]{barchid2022bina}
Sami Barchid, Jos{\'e} Mennesson, and Chaabane Dj{\'e}raba.
\newblock Bina-rep event frames: A simple and effective representation for event-based cameras.
\newblock In \emph{2022 IEEE International Conference on Image Processing (ICIP)}, pages 3998--4002. IEEE, 2022.

\bibitem[Bonazzi et~al.(2024)Bonazzi, Bian, Lippolis, Li, Sheik, and Magno]{bonazzi2024retina}
Pietro Bonazzi, Sizhen Bian, Giovanni Lippolis, Yawei Li, Sadique Sheik, and Michele Magno.
\newblock Retina: Low-power eye tracking with event camera and spiking hardware.
\newblock In \emph{Proceedings of the IEEE/CVF Conference on Computer Vision and Pattern Recognition}, pages 5684--5692, 2024.

\bibitem[Chen et~al.(2024)Chen, Duanmu, and Long]{chen2024large}
Jiadi Chen, Chunjiang Duanmu, and Huanhuan Long.
\newblock Large kernel frequency-enhanced network for efficient single image super-resolution.
\newblock In \emph{Proceedings of the IEEE/CVF Conference on Computer Vision and Pattern Recognition}, pages 6317--6326, 2024.

\bibitem[Chen et~al.(2023)Chen, Wang, Liu, and Gao]{chen20233et}
Qinyu Chen, Zuowen Wang, Shih-Chii Liu, and Chang Gao.
\newblock 3et: Efficient event-based eye tracking using a change-based convlstm network.
\newblock In \emph{2023 IEEE Biomedical Circuits and Systems Conference (BioCAS)}, pages 1--5. IEEE, 2023.

\bibitem[Cheng et~al.(2024)Cheng, Yue, Fang, Gisler, Ren, Fu, Ma, Huang, Xu, Bouhelier, et~al.]{cheng2024memristive}
Bojun Cheng, ZHOU Yue, Yuetong Fang, Raphael Gisler, Hongwei Ren, Haotian Fu, Zelin Ma, Yulong Huang, Renjing Xu, Alexandre Bouhelier, et~al.
\newblock Memristive blinking neuron enabling dense and scalable photonically-linked neural network.
\newblock 2024.

\bibitem[Cho et~al.(2014)Cho, Van~Merri{\"e}nboer, Gulcehre, Bahdanau, Bougares, Schwenk, and Bengio]{cho2014learning}
Kyunghyun Cho, Bart Van~Merri{\"e}nboer, Caglar Gulcehre, Dzmitry Bahdanau, Fethi Bougares, Holger Schwenk, and Yoshua Bengio.
\newblock Learning phrase representations using rnn encoder-decoder for statistical machine translation.
\newblock \emph{arXiv preprint arXiv:1406.1078}, 2014.

\bibitem[Cooley and Tukey(1965)]{cooley1965algorithm}
James~W Cooley and John~W Tukey.
\newblock An algorithm for the machine calculation of complex fourier series.
\newblock \emph{Mathematics of computation}, 19\penalty0 (90):\penalty0 297--301, 1965.

\bibitem[Deng et~al.(2022)Deng, Chen, Liu, and Li]{deng2022voxel}
Yongjian Deng, Hao Chen, Hai Liu, and Youfu Li.
\newblock A voxel graph cnn for object classification with event cameras.
\newblock In \emph{Proceedings of the IEEE/CVF Conference on Computer Vision and Pattern Recognition}, pages 1172--1181, 2022.

\bibitem[Ding et~al.(2024)Ding, Wang, Gao, Liu, and Chen]{ding2024facet}
Junyuan Ding, Ziteng Wang, Chang Gao, Min Liu, and Qinyu Chen.
\newblock Facet: Fast and accurate event-based eye tracking using ellipse modeling for extended reality.
\newblock \emph{arXiv preprint arXiv:2409.15584}, 2024.

\bibitem[Elman(1990)]{elman1990finding}
Jeffrey~L Elman.
\newblock Finding structure in time.
\newblock \emph{Cognitive science}, 14\penalty0 (2):\penalty0 179--211, 1990.

\bibitem[Guestrin and Eizenman(2006)]{guestrin2006general}
Elias~Daniel Guestrin and Moshe Eizenman.
\newblock General theory of remote gaze estimation using the pupil center and corneal reflections.
\newblock \emph{IEEE Transactions on biomedical engineering}, 53\penalty0 (6):\penalty0 1124--1133, 2006.

\bibitem[Hochreiter and Schmidhuber(1997)]{hochreiter1997long}
Sepp Hochreiter and J{\"u}rgen Schmidhuber.
\newblock Long short-term memory.
\newblock \emph{Neural computation}, 9\penalty0 (8):\penalty0 1735--1780, 1997.

\bibitem[Hu et~al.(2021)Hu, Liu, and Delbruck]{hu2021v2e}
Yuhuang Hu, Shih-Chii Liu, and Tobi Delbruck.
\newblock v2e: From video frames to realistic dvs events.
\newblock In \emph{Proceedings of the IEEE/CVF conference on computer vision and pattern recognition}, pages 1312--1321, 2021.

\bibitem[Iddrisu et~al.(2024)Iddrisu, Shariff, Corcoran, O’Connor, Lemley, and Little]{iddrisu2024event}
Khadija Iddrisu, Waseem Shariff, Peter Corcoran, Noel O’Connor, Joe Lemley, and Suzanne Little.
\newblock Event camera based eye motion analysis: A survey.
\newblock \emph{IEEE Access}, 2024.

\bibitem[Jiang et~al.(2024)Jiang, Wang, Yu, and He]{jiang2024eye}
Yizhou Jiang, Wenwei Wang, Lei Yu, and Chu He.
\newblock Eye tracking based on event camera and spiking neural network.
\newblock \emph{Electronics}, 13\penalty0 (14):\penalty0 2879, 2024.

\bibitem[Jin et~al.(2024)Jin, Chai, Tang, Zhou, and Wang]{jin2024eye}
Xin Jin, Suyu Chai, Jie Tang, Xianda Zhou, and Kai Wang.
\newblock Eye-tracking in ar/vr: A technological review and future directions.
\newblock \emph{IEEE Open Journal on Immersive Displays}, 2024.

\bibitem[Kim et~al.(2019)Kim, Stengel, Majercik, De~Mello, Dunn, Laine, McGuire, and Luebke]{kim2019nvgaze}
Joohwan Kim, Michael Stengel, Alexander Majercik, Shalini De~Mello, David Dunn, Samuli Laine, Morgan McGuire, and David Luebke.
\newblock Nvgaze: An anatomically-informed dataset for low-latency, near-eye gaze estimation.
\newblock In \emph{Proceedings of the 2019 CHI conference on human factors in computing systems}, pages 1--12, 2019.

\bibitem[Kim et~al.(2024)Kim, Cho, and Yoon]{kim2024frequency}
Taewoo Kim, Hoonhee Cho, and Kuk-Jin Yoon.
\newblock Frequency-aware event-based video deblurring for real-world motion blur.
\newblock In \emph{Proceedings of the IEEE/CVF Conference on Computer Vision and Pattern Recognition}, pages 24966--24976, 2024.

\bibitem[Konrad et~al.(2020)Konrad, Angelopoulos, and Wetzstein]{konrad2020gaze}
Robert Konrad, Anastasios Angelopoulos, and Gordon Wetzstein.
\newblock Gaze-contingent ocular parallax rendering for virtual reality.
\newblock \emph{ACM Transactions on Graphics (TOG)}, 39\penalty0 (2):\penalty0 1--12, 2020.

\bibitem[Lai et~al.(2014)Lai, Shih, and Hung]{lai2014hybrid}
Chih-Chuan Lai, Sheng-Wen Shih, and Yi-Ping Hung.
\newblock Hybrid method for 3-d gaze tracking using glint and contour features.
\newblock \emph{IEEE Transactions on Circuits and Systems for Video Technology}, 25\penalty0 (1):\penalty0 24--37, 2014.

\bibitem[Li et~al.(2018)Li, You, and Robles-Kelly]{li2018frequency}
Junxuan Li, Shaodi You, and Antonio Robles-Kelly.
\newblock A frequency domain neural network for fast image super-resolution.
\newblock In \emph{2018 International Joint Conference on Neural Networks (IJCNN)}, pages 1--8. IEEE, 2018.

\bibitem[Li et~al.(2023)Li, Bhat, and Raychowdhury]{li2023track}
Nealson Li, Ashwin Bhat, and Arijit Raychowdhury.
\newblock E-track: Eye tracking with event camera for extended reality (xr) applications.
\newblock In \emph{2023 IEEE 5th International Conference on Artificial Intelligence Circuits and Systems (AICAS)}, pages 1--5. IEEE, 2023.

\bibitem[Lichtsteiner et~al.(2008)Lichtsteiner, Posch, and Delbruck]{lichtsteiner2008128}
Patrick Lichtsteiner, Christoph Posch, and Tobi Delbruck.
\newblock A 128{$\times$}128 120 db 15{$\mu$}s latency asynchronous temporal contrast vision sensor.
\newblock \emph{IEEE journal of solid-state circuits}, 43\penalty0 (2):\penalty0 566--576, 2008.

\bibitem[Lin et~al.(2024)Lin, Ren, and Cheng]{lin2024fapnet}
Xiaopeng Lin, Hongwei Ren, and Bojun Cheng.
\newblock Fapnet: An effective frequency adaptive point-based eye tracker.
\newblock In \emph{Proceedings of the IEEE/CVF Conference on Computer Vision and Pattern Recognition}, pages 5789--5798, 2024.

\bibitem[Lu et~al.(2014)Lu, Sugano, Okabe, and Sato]{lu2014adaptive}
Feng Lu, Yusuke Sugano, Takahiro Okabe, and Yoichi Sato.
\newblock Adaptive linear regression for appearance-based gaze estimation.
\newblock \emph{IEEE transactions on pattern analysis and machine intelligence}, 36\penalty0 (10):\penalty0 2033--2046, 2014.

\bibitem[Ma et~al.(2024)Ma, Yuan, Xie, Ren, Liu, He, Ren, Yu, and Ni]{ma2024generative}
Fei Ma, Yucheng Yuan, Yifan Xie, Hongwei Ren, Ivan Liu, Ying He, Fuji Ren, Fei~Richard Yu, and Shiguang Ni.
\newblock Generative technology for human emotion recognition: A scoping review.
\newblock \emph{Information Fusion}, page 102753, 2024.

\bibitem[Ma et~al.(2022)Ma, Qin, You, Ran, and Fu]{ma2022rethinking}
Xu Ma, Can Qin, Haoxuan You, Haoxi Ran, and Yun Fu.
\newblock Rethinking network design and local geometry in point cloud: A simple residual mlp framework.
\newblock \emph{arXiv preprint arXiv:2202.07123}, 2022.

\bibitem[Mao et~al.(2023)Mao, Liu, Liu, Li, Shen, and Wang]{mao2023intriguing}
Xintian Mao, Yiming Liu, Fengze Liu, Qingli Li, Wei Shen, and Yan Wang.
\newblock Intriguing findings of frequency selection for image deblurring.
\newblock In \emph{Proceedings of the AAAI Conference on Artificial Intelligence}, pages 1905--1913, 2023.

\bibitem[Mazzeo et~al.(2021)Mazzeo, D'Amico, Spagnolo, and Distante]{mazzeo2021deep}
Pier~Luigi Mazzeo, Dilan D'Amico, Paolo Spagnolo, and Cosimo Distante.
\newblock Deep learning based eye gaze estimation and prediction.
\newblock In \emph{2021 6th International Conference on Smart and Sustainable Technologies (SpliTech)}, pages 1--6. IEEE, 2021.

\bibitem[Mestre et~al.(2018)Mestre, Gautier, and Pujol]{mestre2018robust}
Clara Mestre, Josselin Gautier, and Jaume Pujol.
\newblock Robust eye tracking based on multiple corneal reflections for clinical applications.
\newblock \emph{Journal of biomedical optics}, 23\penalty0 (3):\penalty0 035001--035001, 2018.

\bibitem[Morimoto and Mimica(2005)]{morimoto2005eye}
Carlos~H Morimoto and Marcio~RM Mimica.
\newblock Eye gaze tracking techniques for interactive applications.
\newblock \emph{Computer vision and image understanding}, 98\penalty0 (1):\penalty0 4--24, 2005.

\bibitem[Newell et~al.(2016)Newell, Yang, and Deng]{newell2016stacked}
Alejandro Newell, Kaiyu Yang, and Jia Deng.
\newblock Stacked hourglass networks for human pose estimation.
\newblock In \emph{Computer Vision--ECCV 2016: 14th European Conference, Amsterdam, The Netherlands, October 11-14, 2016, Proceedings, Part VIII 14}, pages 483--499. Springer, 2016.

\bibitem[Pei et~al.(2024)Pei, Br{\"u}ers, Crouzet, McLelland, and Coenen]{pei2024lightweight}
Yan~Ru Pei, Sasskia Br{\"u}ers, S{\'e}bastien Crouzet, Douglas McLelland, and Olivier Coenen.
\newblock A lightweight spatiotemporal network for online eye tracking with event camera.
\newblock In \emph{Proceedings of the IEEE/CVF Conference on Computer Vision and Pattern Recognition}, pages 5780--5788, 2024.

\bibitem[Qi et~al.(2017{\natexlab{a}})Qi, Su, Mo, and Guibas]{qi2017pointnet}
Charles~R Qi, Hao Su, Kaichun Mo, and Leonidas~J Guibas.
\newblock Pointnet: Deep learning on point sets for 3d classification and segmentation.
\newblock In \emph{Proceedings of the IEEE conference on computer vision and pattern recognition}, pages 652--660, 2017{\natexlab{a}}.

\bibitem[Qi et~al.(2017{\natexlab{b}})Qi, Yi, Su, and Guibas]{qi2017pointnet++}
Charles~Ruizhongtai Qi, Li Yi, Hao Su, and Leonidas~J Guibas.
\newblock Pointnet++: Deep hierarchical feature learning on point sets in a metric space.
\newblock \emph{Advances in neural information processing systems}, 30, 2017{\natexlab{b}}.

\bibitem[Ramachandran(2002)]{ramachandran2002encyclopedia}
Vilayanur~S Ramachandran.
\newblock \emph{Encyclopedia of the human brain}.
\newblock Elsevier, 2002.

\bibitem[Rebecq et~al.(2019)Rebecq, Ranftl, Koltun, and Scaramuzza]{rebecq2019high}
Henri Rebecq, Ren{\'e} Ranftl, Vladlen Koltun, and Davide Scaramuzza.
\newblock High speed and high dynamic range video with an event camera.
\newblock \emph{IEEE transactions on pattern analysis and machine intelligence}, 43\penalty0 (6):\penalty0 1964--1980, 2019.

\bibitem[Ren et~al.(2023)Ren, Zhou, Huang, Fu, Lin, Song, and Cheng]{ren2023spikepoint}
Hongwei Ren, Yue Zhou, Yulong Huang, Haotian Fu, Xiaopeng Lin, Jie Song, and Bojun Cheng.
\newblock Spikepoint: An efficient point-based spiking neural network for event cameras action recognition.
\newblock \emph{arXiv preprint arXiv:2310.07189}, 2023.

\bibitem[Ren et~al.(2024{\natexlab{a}})Ren, Zhou, Zhu, Fu, Huang, Lin, Fang, Ma, Yu, and Cheng]{ren2024rethinking}
Hongwei Ren, Yue Zhou, Jiadong Zhu, Haotian Fu, Yulong Huang, Xiaopeng Lin, Yuetong Fang, Fei Ma, Hao Yu, and Bojun Cheng.
\newblock Rethinking efficient and effective point-based networks for event camera classification and regression: Eventmamba.
\newblock \emph{arXiv preprint arXiv:2405.06116}, 2024{\natexlab{a}}.

\bibitem[Ren et~al.(2024{\natexlab{b}})Ren, Zhu, Zhou, Fu, Huang, and Cheng]{ren2024simple}
Hongwei Ren, Jiadong Zhu, Yue Zhou, Haotian Fu, Yulong Huang, and Bojun Cheng.
\newblock A simple and effective point-based network for event camera 6-dofs pose relocalization.
\newblock \emph{arXiv preprint arXiv:2403.19412}, 2024{\natexlab{b}}.

\bibitem[Ryan et~al.(2023)Ryan, Elrasad, Shariff, Lemley, Kielty, Hurney, and Corcoran]{ryan2023real}
Cian Ryan, Amr Elrasad, Waseem Shariff, Joe Lemley, Paul Kielty, Patrick Hurney, and Peter Corcoran.
\newblock Real-time multi-task facial analytics with event cameras.
\newblock \emph{IEEE Access}, 11:\penalty0 76964--76976, 2023.

\bibitem[Shariff et~al.(2024)Shariff, Dilmaghani, Kielty, Moustafa, Lemley, and Corcoran]{shariff2024event}
Waseem Shariff, Mehdi~Sefidgar Dilmaghani, Paul Kielty, Mohamed Moustafa, Joe Lemley, and Peter Corcoran.
\newblock Event cameras in automotive sensing: A review.
\newblock \emph{IEEE Access}, 2024.

\bibitem[Stein et~al.(2021)Stein, Niehorster, Watson, Steinicke, Rifai, Wahl, and Lappe]{stein2021comparison}
Niklas Stein, Diederick~C Niehorster, Tamara Watson, Frank Steinicke, Katharina Rifai, Siegfried Wahl, and Markus Lappe.
\newblock A comparison of eye tracking latencies among several commercial head-mounted displays.
\newblock \emph{i-Perception}, 12\penalty0 (1):\penalty0 2041669520983338, 2021.

\bibitem[Wang and Ji(2017)]{wang2017real}
Kang Wang and Qiang Ji.
\newblock Real time eye gaze tracking with 3d deformable eye-face model.
\newblock In \emph{Proceedings of the IEEE International Conference on Computer Vision}, pages 1003--1011, 2017.

\bibitem[Wang et~al.(2024{\natexlab{a}})Wang, Gao, Wu, Conde, Timofte, Liu, Chen, Zha, Zhai, Han, et~al.]{wang2024event}
Zuowen Wang, Chang Gao, Zongwei Wu, Marcos~V Conde, Radu Timofte, Shih-Chii Liu, Qinyu Chen, Zheng-jun Zha, Wei Zhai, Han Han, et~al.
\newblock Event-based eye tracking. ais 2024 challenge survey.
\newblock In \emph{Proceedings of the IEEE/CVF Conference on Computer Vision and Pattern Recognition}, pages 5810--5825, 2024{\natexlab{a}}.

\bibitem[Wang et~al.(2024{\natexlab{b}})Wang, Wan, Han, Liao, Wu, Zhai, Cao, and Zha]{wang2024mambapupil}
Zhong Wang, Zengyu Wan, Han Han, Bohao Liao, Yuliang Wu, Wei Zhai, Yang Cao, and Zheng-Jun Zha.
\newblock Mambapupil: Bidirectional selective recurrent model for event-based eye tracking.
\newblock In \emph{Proceedings of the IEEE/CVF Conference on Computer Vision and Pattern Recognition}, pages 5762--5770, 2024{\natexlab{b}}.

\bibitem[Wei et~al.(2016)Wei, Ramakrishna, Kanade, and Sheikh]{wei2016convolutional}
Shih-En Wei, Varun Ramakrishna, Takeo Kanade, and Yaser Sheikh.
\newblock Convolutional pose machines.
\newblock In \emph{Proceedings of the IEEE conference on Computer Vision and Pattern Recognition}, pages 4724--4732, 2016.

\bibitem[Yang et~al.(2023)Yang, Liu, Duan, Fan, Zhang, and Jin]{yang2023three}
Ziyi Yang, Kehan Liu, Yiru Duan, Mingjia Fan, Qiyue Zhang, and Zhou Jin.
\newblock Three challenges in reram-based process-in-memory for neural network.
\newblock In \emph{2023 IEEE 5th International Conference on Artificial Intelligence Circuits and Systems (AICAS)}, pages 1--5. IEEE, 2023.

\bibitem[Zemblys et~al.(2019)Zemblys, Niehorster, and Holmqvist]{zemblys2019gazenet}
Raimondas Zemblys, Diederick~C Niehorster, and Kenneth Holmqvist.
\newblock gazenet: End-to-end eye-movement event detection with deep neural networks.
\newblock \emph{Behavior research methods}, 51:\penalty0 840--864, 2019.

\bibitem[Zhao et~al.(2023)Zhao, Zheng, Liu, Wang, and Chen]{zhao2023poseformerv2}
Qitao Zhao, Ce Zheng, Mengyuan Liu, Pichao Wang, and Chen Chen.
\newblock Poseformerv2: Exploring frequency domain for efficient and robust 3d human pose estimation.
\newblock In \emph{Proceedings of the IEEE/CVF conference on computer vision and pattern recognition}, pages 8877--8886, 2023.

\end{thebibliography}
}


\end{document}